\documentclass[a4paper,conference]{IEEEtran}

\usepackage{ctable}
\usepackage{xcolor}
\usepackage{booktabs}
\usepackage{adjustbox}
\usepackage{multirow}
\usepackage{pifont}
\usepackage{amsmath}
\DeclareMathOperator*{\argmax}{arg\,max}
\DeclareMathOperator*{\argmin}{arg\,min}
\usepackage{amssymb}
\usepackage{pifont}
\usepackage{graphicx}

\usepackage{pifont}
\usepackage{amsmath}
\usepackage{amssymb}
\usepackage{color,soul}

\newcommand{\mytilde}{\raise.17ex\hbox{$\scriptstyle\mathtt{\sim}$}}

\def\ll{\mathbf{l}}

\def\xx{\mathbf{x}}

\def\AA{\mathbf{A}}

\def\CC{\mathbf{C}}
\def\DD{\mathbf{D}}

\usepackage[hidelinks]{hyperref}
\usepackage{cleveref}
\hypersetup{
  colorlinks   = true, 
  urlcolor     = blue, 
  linkcolor    = blue, 
  citecolor   = red 
}

\begin{document}

\title{Effects of Auxiliary Knowledge on Continual Learning}

\author{\IEEEauthorblockN{Giovanni Bellitto, Matteo Pennisi, \\ Simone Palazzo, Concetto Spampinato}
\IEEEauthorblockA{PeRCeiVe Lab, \\ University of Catania, Italy.\\ 
Email: name.surname@unict.it}
\and
\IEEEauthorblockN{Lorenzo Bonicelli, Matteo Boschini, \\ Simone Calderara}
\IEEEauthorblockA{AImageLab, \\  University of Modena and Reggio Emilia.\\ 
Email: name.surname@unimore.it}

}

\maketitle

\begin{abstract}
In Continual Learning (CL), a neural network is trained on a stream of data whose distribution changes over time. In this context, the main problem is how to learn new information without forgetting old knowledge (i.e., Catastrophic Forgetting). Most existing CL approaches focus on finding solutions to preserve acquired knowledge, so working on the “past” of the model. However, we argue that as the model has to continually learn new tasks, it is also important to put focus on the “present” knowledge that could improve following tasks learning. In this paper we propose a new, simple, CL algorithm that focuses on solving the current task in a way that might facilitate the learning of the next ones. More specifically, our approach combines the main data stream with a secondary, diverse and uncorrelated  stream, from which the network can draw auxiliary knowledge. This helps the model from different perspectives, since auxiliary data may contain useful features for the current and the next tasks and incoming task classes can be mapped onto auxiliary classes. Furthermore, the addition of data to the current task is implicitly making the classifier more robust as we are forcing the extraction of more discriminative features. Our method can outperform existing state-of-the-art models on the most common CL Image Classification benchmarks.

\end{abstract}

\IEEEpeerreviewmaketitle

\section{Introduction}
Human beings and animals are naturally able to memorize information presented in a sequence~\cite{cichon2015branch}; on the contrary, Artificial Neural Networks (ANNs) learning from a non-i.i.d.\ stream of data incur in \textit{Catastrophic Forgetting}~\cite{mccloskey1989catastrophic, ratcliff1990connectionist}. Continual Learning (CL)~\cite{parisi2019continual, delange2021continual} aims at designing methods that compensate for this issue and facilitate the retention of previous knowledge either by means of regularization~\cite{kirkpatrick2017overcoming, li2017learning}, architectural designs~\cite{rusu2016progressive, serra2018overcoming} or (pseudo-)replay of past data~\cite{ratcliff1990connectionist, chaudhry2019tiny, shin2017continual}.

The insurgence of catastrophic forgetting is ascribed to the tendency of models to rewrite their hidden representations as they adjust their parameters to best fit an input distribution that changes in time~\cite{french1991using}. However, \emph{McRae \& Hetherington} 
highlight a meaningful difference in the way humans and ML models learn from a sequence of data: whenever human subjects are evaluated on their ability to memorize a sequence of concepts, they start out possessing an already-large body of knowledge~\cite{mcrae1993catastrophic}.
In other words, humans are generalists that can anchor novel data in the context of previous knowledge, while ANNs must specialize on a limited pool of data at each time without any additional reference.

An obvious choice to bridge this gap is \textit{pre-training} the models on a large amount of available off-the-shelf i.i.d.\ data, leading to a better initialization for the learning procedure~\cite{mcrae1993catastrophic, hendrycks2019using}. However, we observe that pre-training is not always rewarding in a CL setting, especially in case of small-size replay memories: the ever-changing stream of data entails large changes in model parameters, leading to the forgetting of the pre-training.

We instead propose a learning strategy to limit catastrophic forgetting by providing an additional data stream (uncorrelated from the target data), from which the network can draw auxiliary knowledge. The role of this data stream is to provide models with a more stable  representation of the world that can be re-used for incrementally learning new classes or categories leveraging the already-learned low-level features. 
Indeed, it appears that the human brain can adapt and rewire itself more easily when learning new things related to familiar skills because pre-existing neuronal structure constrains what one can learn~\cite{ddf607f82ffd4ff4a2e7e9e4d833a78d}. We attempt to enforce this concept into CL through the definition of \emph{an associative rule} that helps learning new classes by measuring the simultaneous firing of neurons between past knowledge and the current data stream. This is implemented through a simple yet effective strategy named \emph{MAH}, that, during a new task, assigns new classes to model's corresponding \textbf{m}ost \textbf{a}ctivated \textbf{h}eads.

Experimental results carried out on standard CL settings, involving CIFAR-10 and (a subset of) Tiny-ImageNet benchmarks, demonstrate that using a separate auxiliary data stream is mostly beneficial with limited size buffer leading to a performance gain of several percent points w.r.t.\ state-of-the-art methods. Analogously, the MAH strategy reveals to be more effective than the standard class mapping procedure independently from the buffer size. We also investigate the role of model pre-training as compared to sustained auxiliary data employment highlighting  that, for small-size buffers, auxiliary data is to be preferred to pre-training.

\begin{figure*}[h!]
\centering
\includegraphics[width=0.95\textwidth]{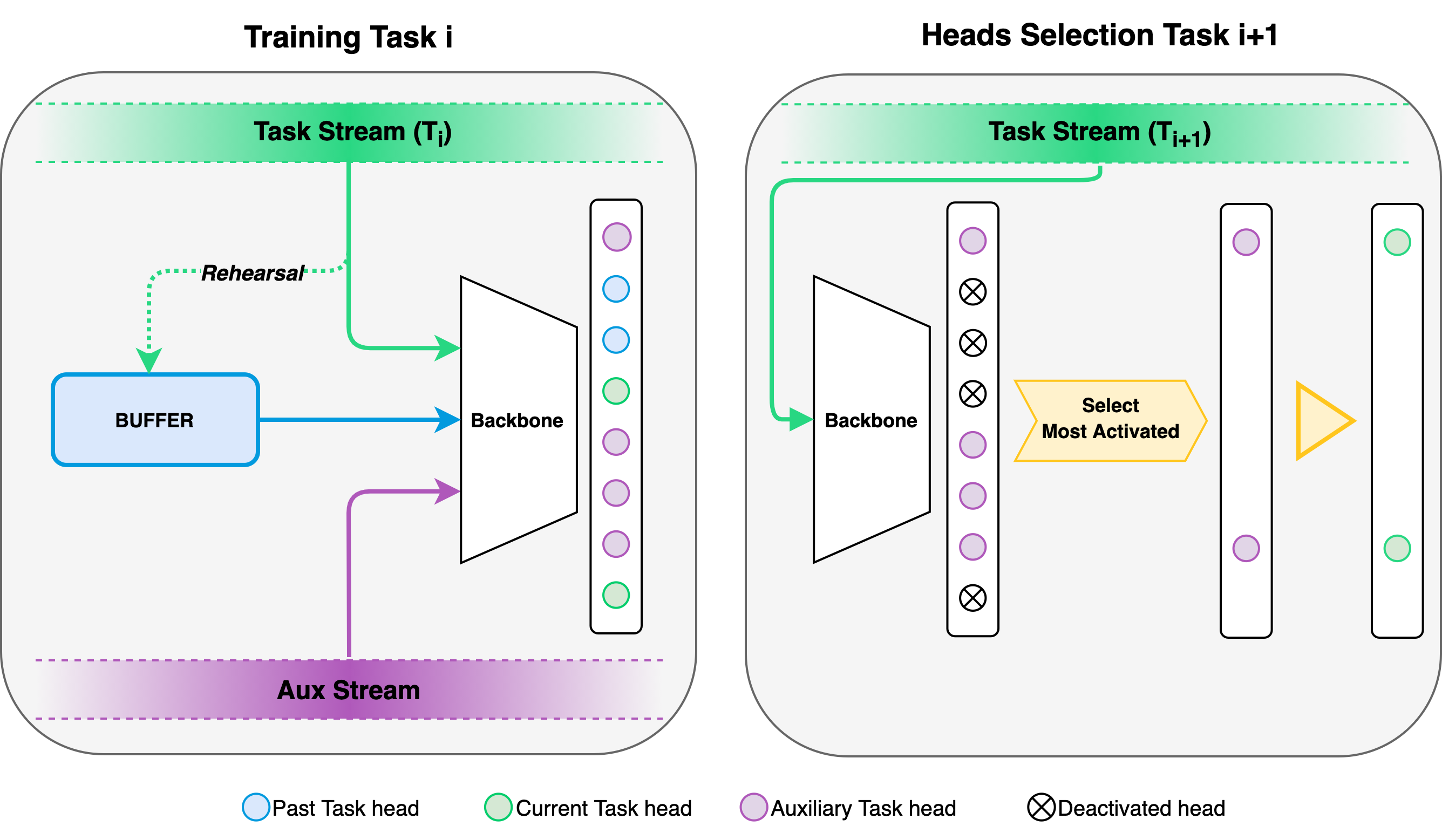}
\caption{During the training of the current Task $T_i$ the model is trained combining the ``current'' data coming from the Task Stream and the ``past'' data stored inside the buffer. In addition, the remaining heads are trained using the auxiliary data stream. At the beginning of a new Task $T_{i+1}$ the MAH procedure is conducted as follows: 1) Only the heads trained with the auxiliary data are kept activated; 2) the $T_{i+1}$ task is forwarded to the frozen model in order to store activation information about the heads; 3) for each class in task $T_{i+1}$, each new class is assigned to
the head that activates the most, replacing the corresponding auxiliary data class.}
\label{fig:architecture}
\end{figure*}

In conclusion, our strategy is beneficial in Continual Learning from multiple perspectives: the model avoids overfitting current examples, learns more general features and -- as auxiliary data-points stand in for future examples -- better prepares to learn future classes by suitably associating past knowledge to the new acquired one. All these aspects are mainly observed with reduced buffer size, thus contributing to the efforts that aim at generalizing CL approaches to real-world scenarios.

\section{Related Work}

The seminal study by \textit{McCloskey and Cohen} first drew attention to the tendency of ANNs to forget previously learned knowledge catastrophically~\cite{mccloskey1989catastrophic}. In spite of the outstanding results achieved by deep learning models in recent years~\cite{silver2016mastering, vinyals2019grandmaster}, this problem still persists and prevents ANNs from learning flexibly from non-i.i.d.\ data-streams. To tackle this issue, researchers and practitioners design CL methods, i.e., strategies that make machine learning models retain high accuracy on previously seen data when trained on an ever-changing input distribution~\cite{parisi2019continual, delange2021continual}. While many distinct strategies have been applied for this purpose, CL approaches can be broadly categorized into two families: \textit{structuring} or \textit{generalization}.

\subsection{Structuring approaches to Continual Learning}
Methods in the first class aim at making interference between distinct concepts less likely by endowing the stored knowledge with a disentangled structure.
\cite{lewandowsky1995catastrophic} first pioneered the idea to reduce forgetting by orthogonalizing feature representations. A similar approach was recently taken by~\cite{chaudhry2020continual}. Alternatively, \textit{structuring} can be pursued at an architectural level, by explicitly allocating distinct subsets of model parameters to distinct tasks~\cite{rusu2016progressive, mallya2018packnet, jung2020continual}, encouraging non-overlapping activation patterns for different data~\cite{serra2018overcoming, rajasegaran2019random}, or by simply applying dropout~\cite{goodfellow2014empirical, mirzadeh2020understanding}. Finally, several approaches regularize back-propagation by projecting the gradient to minimize the interference between tasks learned at different times~\cite{lopez2017gradient, chaudhry2018efficient, farajtabar2020orthogonal}. While \textit{structuring} approaches are usually characterized by a simpler training procedure, they typically require the availability of a task-identifier at test time.

\subsection{Generalization approaches to Continual Learning}
At the opposite end of the spectrum, \textit{generalization} methods prevent forgetting by encouraging the model to compare and contrast input data encountered throughout the sequence, thus recovering the i.i.d.\ property of training~\cite{mcrae1993catastrophic}. Most notably, rehearsal-based approaches do so by maintaining a working memory of previously seen examples and interleaving them with the input data~\cite{ratcliff1990connectionist, chaudhry2019tiny, buzzega2020dark, aljundi2019gradient, aljundi2019online}, while pseudo-rehearsal methods approximate this procedure with a generative model~\cite{riemer2019scalable, shin2017continual}. Other works prioritize the learning of high-level representations either by adopting learning objectives designed not to disrupt the performance on previous tasks~\cite{rebuffi2017icarl, hou2019learning, li2017learning, wu2019large}, or by making use of semi-supervised learning techniques to learn general features~\cite{cha2021co2l, pham2021dualnet, kim2021continual}. The \textit{generalization} approach is taken to an extreme by \cite{prabhu2020gdumb}, which shows satisfactory results on CL benchmarks by training a model in an i.i.d.\ fashion on samples gathered greedily from the input stream. 

\textit{Generalization} strategies naturally blend knowledge gathered at different times to build a unified predictor, making them more reliable than their \textit{structuring} counterparts in the realistic settings where no task-identifier is given at testing-time~\cite{farquhar2018towards, van2019three}. The approach proposed in this paper aligns with the former group of methods; indeed, we argue that generalization should be extended beyond already-seen data and embrace yet-unseen knowledge as well. 

\begin{table*}[ht!]
    \centering
    \caption{Accuracy on Split-CIFAR-10 for several Continual Learning methods. Best results in bold, second-best in italic.}
\begin{adjustbox}{width=\textwidth} 
\begin{tabular}{clcccccccccc}
\toprule
\multirow{5}{*}{\rotatebox[origin=c]{90}{\textit{Class-IL}}} & \textit{Buffer}    &           ER~\cite{riemer2018learning} &          GEM~\cite{lopez2017gradient} &        A-GEM~\cite{chaudhry2018efficient} &        iCaRL~\cite{rebuffi2017icarl} &          FDR~\cite{benjamin2018measuring} &          GSS~\cite{aljundi2019gradient} &           HAL~\cite{chaudhry2020using} &           DER~\cite{buzzega2020dark} &        DER++~\cite{buzzega2020dark} &  \textbf{Ours} \\ 
\cmidrule{2-12}
& 50          &    32.69 ± 0.39 & 22.10 ± 0.41 &   20.02 ± 0.08  &    \textit{55.51 ± 1.64} & 28.32 ± 4.51 &   26.62 ± 1.36 & 25.26 ± 1.73 & 44.85 ± 2.71 & 49.28 ± 3.16 & \textbf{56.33 ± 0.95}   \\
& 200         & 44.79 ± 1.86 & 25.54 ± 0.76 & 20.04 ± 0.34 & 49.02 ± 3.20 & 30.91 ± 2.74 & 39.07 ± 5.59 & 32.36 ± 2.70  & 61.93 ± 1.79 & \textit{64.88 ± 1.17} & \textbf{70.86 ± 0.95}   \\
& 500         & 57.74 ± 0.27 & 26.20 ± 1.26 & 22.67 ± 0.57 & 47.55 ± 3.95 & 28.71 ± 3.23 & 49.73 ± 4.78 & 41.79 ± 4.46  & 70.51 ± 1.67 & \textit{72.70 ± 1.36} & \textbf{75.07 ± 0.41}   \\
& 5120        & 82.47 ± 0.52 & 25.56 ± 3.46 & 21.99 ± 2.29 & 55.07 ± 1.55 & 19.70 ± 0.07 & 67.27 ± 4.27 & 59.12 ± 4.41  & 83.81 ± 0.33 & \textbf{85.24 ± 0.49} & \textit{84.56 ± 0.55}   \\
\midrule

\multirow{5}{*}{\rotatebox[origin=c]{90}{\textit{Task-IL}}} & \textit{Buffer}    &           ER~\cite{riemer2018learning} &          GEM~\cite{lopez2017gradient} &        A-GEM~\cite{chaudhry2018efficient} &        iCaRL~\cite{rebuffi2017icarl} &          FDR~\cite{benjamin2018measuring} &          GSS~\cite{aljundi2019gradient} &           HAL~\cite{chaudhry2020using} &           DER~\cite{buzzega2020dark} &        DER++~\cite{buzzega2020dark} &  \textbf{Ours} \\
\cmidrule{2-12}
& 50          & 86.98 ± 1.19 & 81.36 ± 1.43 &     81.09 ± 1.88 &  \textit{88.86} ± 2.51 & 85.23 ± 1.24 &  85.22 ± 1.03 & 78.73 ± 3.16  & 85.04 ± 1.17 & 86.14 ± 2.56 & \textbf{89.57 ± 2.47}   \\
& 200         & 91.19 ± 0.94 & 90.44 ± 0.94 & 83.88 ± 1.49 & 88.99 ± 2.13 & 91.01 ± 0.68 & 88.80 ± 2.89 & 82.51 ± 3.20  & 91.40 ± 0.92 & \textit{91.92 ± 0.60} & \textbf{93.30 ± 0.64}   \\
& 500         & 93.61 ± 0.27 & 92.16 ± 0.69 & 89.48 ± 1.45 & 88.22 ± 2.62 & 93.29 ± 0.59 & 91.02 ± 1.57 & 84.54 ± 2.36  & 93.40 ± 0.39 & \textbf{93.88 ± 0.50} & \textit{93.62 ± 0.58}   \\
& 5120        & \textbf{96.98 ± 0.17} & 95.55 ± 0.02 & 90.10 ± 2.09 & 92.23 ± 0.84 & 94.32 ± 0.97 & 94.19 ± 1.15 & 88.51 ± 3.32  & 95.43 ± 0.33 & \textit{96.12 ± 0.21} & 95.84 ± 0.42   \\
\bottomrule
\end{tabular}
\end{adjustbox}
    
    \label{tab:seqCIFAR10}
\end{table*}

\section{Method}
\label{sec:method}

Most CL methods use current and, if the method implies a rehearsal strategy, past task classification heads during the training of the current task. Future heads, that will be mapped to classes from following tasks, are not involved in the process at all. This poses a potentially dangerous situation due to the model minimizing its prediction scores for future heads, which results in a high loss peak when these heads are used at the beginning of future tasks.

We propose to leverage an auxiliary data stream, not correlated with the main task stream, in order to keep these future heads activated since the beginning of training. The proposed strategy is also beneficial to learn more distinguishing and reusable features, as the model cannot focus on simply discriminating between the classes from the task at hand. Furthermore, since auxiliary training leads future task heads to learn to recognize their own specific patterns, we exploit this property to devise a ``most activated heads'' (MAH) assignment strategy for future classes, that minimizes the loss peak that the model typically incurs at the beginning of a new task. Hence, the use of an auxiliary stream favors the current task and improves forward transfer to future tasks. The proposed approach is illustrated in Fig.~\ref{fig:architecture}.

Formally, a typical CL classification problem requires solving several tasks sequentially, where each task $T_t$, with $t\in \left\{1, ..., T \right\}$ and $T$ being the number of tasks, consists in learning to classify a set of classes $\CC_t$. In this work, we follow the common Class-IL and Task-IL settings~\cite{van2019three}, which assume no overlap between classes from different tasks.

Each task is associated with an i.i.d.\ distribution $D_t$ of $(\xx,y)$ pairs of a data point with the corresponding class label from $\CC_t$. In practice, the distribution is approximated by a finite set of samples, i.e., $\DD_t = \left\{ \left(\xx_1, y_1\right) , \left(\xx_2, y_2\right), \dots, \left(\xx_{N_t}, y_{N_t}\right) \right\}$, where $N_t$ is the number of examples for task $t$.

The objective of CL is to find a function $f_\theta$, depending on a set of learnable parameters $\theta$, that minimizes a classification objective over the entire task sequence, such as:
\begin{equation}
 \argmin_\theta \sum_{t=1}^T \sum_{(\xx_i, y_i) \in \DD_t} \mathcal{L}_\text{C}\left( f_\theta(\xx_i), y_i \right),
\end{equation}
where $\mathcal{L}_\text{C}$ is the classification loss (e.g., cross-entropy).

While training for the current task, most recent CL approaches~\cite{kirkpatrick2017overcoming, li2017learning, chaudhry2019tiny, buzzega2020dark} attempt to reduce forgetting by adding an additional loss term that attempts to retain accuracy on previously-seen tasks. The in-task objective at task $t$ then becomes:
\begin{equation}
 \argmin_\theta
 \sum_{(\xx_i, y_i) \in \DD_t} \mathcal{L}_\text{C}\left( f_\theta(\xx_i), y_i \right) +
 \mathcal{L}_\text{CL},
\end{equation}

where $\mathcal{L}_\text{CL}$ is a generic additional loss term that implements countermeasures to catastrophic forgetting (e.g., experience replay from a buffer of samples from previous tasks).

In the proposed scenario, an additional distribution of i.i.d.\ auxiliary data $A$, where $A \ne D_t~\forall t$, is available to the model at training time. Again, the distribution is represented by a set of sample/label pairs $\AA = \left\{ \left(\xx_1, y_1\right) , \left(\xx_2, y_2\right), \dots, \left(\xx_{N_t}, y_{N_t}\right) \right\}$, where labels belong to the class set $\CC_A$.

In the following, we explain the two key aspects of the proposed approach: head pre-activation and ``most activated heads'' class mapping.

\subsection{Head pre-activation}

To ensure that the model employs all of its classification heads from the start, we use classes from the auxiliary dataset as ``place-holders'' for classes from future tasks.

The basic requirement of an auxiliary dataset $\AA$ is related to the cardinality of its set of classes, $\left|\CC_A\right|$, which should satisfy the following condition:
\begin{equation}
\left|\CC_A\right| \geq \sum_{t=2}^T \left|\CC_t\right|.
\end{equation}
In other words, the number of auxiliary classes should be at least equal to the total number of classes in the sequence of continual learning tasks, minus the number of classes from the first task. This guarantees that, when training on the first task, the auxiliary dataset provides enough classes for the classification heads reserved to future tasks.

\emph{Before} starting to train on the first task $t=1$, we randomly choose a subset $\CC_{A,t} \subseteq \CC_A$, with cardinality $\left| \CC_{A,t} \right| = \sum_{t=2}^T \left|\CC_t\right|$. Samples from the selected classes are included in the auxiliary sub-dataset $\AA_t$ and class indexes from $\CC_{A,t}$ are re-mapped to the indexes of classes in $\cup_{t=2}^T \CC_t$ corresponding to future tasks.

At task $t$, we merge the corresponding dataset $\DD_t$ and the auxiliary sub-dataset $\AA_t$ and train on the joint set of classes, in order to minimize the following new in-task objective:
\begin{equation}
 \argmin_\theta
 \sum_{(\xx_i, y_i) \in \DD_t \cup \AA_t} \mathcal{L}_\text{C}\left( f_\theta(\xx_i), y_i \right) + 
 \mathcal{L}_\text{CL}.
 \label{eq:aux}
\end{equation}

As a result, we ensure that all classification heads are employed, reducing the risk of loss peaks on new tasks, and encourage the model to learn more complex, discriminative and stable features.

\subsection{``Most activated heads'' (MAH) class mapping}

At the beginning of each task $t > 1$, it is necessary to update the set of auxiliary classes in $\AA_t$, since $\left| \CC_t \right|$ classes must be removed to make room for classes from the new task.

Moreover, in this scenario, it also makes sense to \emph{assign} the specific heads that will correspond to classes in the new task, rather than simply associating them to the next available heads. An appropriate class mapping can make better (re)use of features learned by the model for classification of auxiliary classes, and reduce high losses that may lead to forgetting previously-learned features.

Our head assignment approach, named MAH from ``most activated heads'', acts before beginning to train on task $t > 1$, by first computing the average logits $\ll_c$, i.e., pre-softmax head activations, for each task class $c \in \CC_t$: to this aim, we select the subset $\DD_{t,c} \subset \DD_t$ which only contains elements of class $c$, and average the corresponding logit vectors as returned by model $f_\theta$:
\begin{equation}
\ll_c = \frac{1}{N_{t,c}} \sum_{\left(\xx_i,y_i\right) \in \DD_{t,c}} f_\theta\left(\xx_i\right),
\end{equation}
where $N_{t,c}$ is the number of elements of class $c$ in $\DD_t$.

Then, each new class in $\CC_t$ is simply associated to the classification head that maximizes its predicted score, i.e., $\argmax \ll_c$. In case of index collisions, largest values are given priority. Finally, the new auxiliary sub-dataset $\AA_t$ is updated by removing from $\AA_{t-1}$ the set of classes corresponding to selected indexes, i.e., $\left\{ \argmax \ll_c \right\}_{c \in \CC_t}$, and training proceeds as previously described.

\section{Experimental Settings}

\subsection{Datasets}

We focus our experiments on two common evaluation protocols~\cite{van2019three}: class-incremental (Class-IL), where the model is asked to gradually solve the complete problem but classes become available at different times; task-incremental (Task-IL), where the model is guided by the task-identity and can only focus to solve each task independently. Specifically, we leverage \textbf{Split-CIFAR-10}~\cite{zenke2017continual}, a widely-used image classification dataset obtained by splitting the 32$\times$32 images of CIFAR-10 into 5 binary tasks. For a more comprehensive evaluation, we also test on the larger 64$\times$64 \textbf{Split-Micro-ImageNet}: a novel benchmark composed of a 20-class subset of Tiny-ImageNet~\cite{Le2015TinyIV}, split into 5 tasks of 4 classes each. 

As for the choice of the auxiliary data, we pair the original data with similarly-sized datasets. In particular, the auxiliary dataset for Split-CIFAR-10 consists of a subset of 10 super-classes from CIFAR-100, selected among those which are not semantic-related to those contained in CIFAR-10. For Split-Micro-ImageNet, we select a subset of 20 classes from ILSVRC-2012, making sure that the chosen data is as unrelated as possible with the original Tiny-ImageNet classes. In detail, 
we first remove Tiny-ImageNet classes from the entire label set; then, we group the remaining 800 classes into 293 super-classes, corresponding to synsets found at distance 8 from the \emph{entity} root node. Finally, we apply Spectral Clustering to select the 20 classes which are most representative of the super-classes.

\subsection{Training procedure}

We apply the approach described in Section~\ref{sec:method} by adapting the DER++~\cite{buzzega2020dark} method, a recent rehearsal-based approach inspired by knowledge distillation principles. For a fair comparison among different models, in our experiments we follow~\cite{buzzega2020dark} and adopt the same training settings. As backbone, we use ResNet-18~\cite{he2016deep} (not pretrained).
We optimize our model with SGD, for 50 epochs on Split-CIFAR-10 and 100 on Split-Micro-ImageNet. During training, samples from the current task and from auxiliary classes are combined so that each mini-batch contains data from both domains. We apply random crops and horizontal flips as data augmentation. All hyperparameters are as defined in~\cite{buzzega2020dark}.

\subsection{Results}
\label{sec:results}

To validate the effectiveness of our approach using auxiliary data during training, we compare our method with other CL methods based on rehearsal strategies: ER~\cite{riemer2018learning},
GEM~\cite{lopez2017gradient}, A-GEM~\cite{chaudhry2018efficient}, iCaRL~\cite{rebuffi2017icarl}, FDR~\cite{benjamin2018measuring}, GSS~\cite{aljundi2019gradient}, HAL~\cite{chaudhry2020using},
DER~\cite{buzzega2020dark} and vanilla DER++~\cite{buzzega2020dark}. Performance for these methods is reported from~\cite{buzzega2020dark}, except for the setup with buffer size equal to 50\footnote{In this case, results were computed using the Mammoth framework for PyTorch: \url{https://github.com/aimagelab/mammoth}}.

As performance metrics, we report classification accuracy in the Class-IL setting as the average of per-task classification accuracy computed after the final task, and in the Task-IL setting as the average of per-task classification accuracy computed at the end of each task.
\begin{table}[h!]
    \centering
\caption{Accuracy on Split-Micro-ImageNet for Experience Replay-based methods.}
\begin{tabular}{llcc}
\toprule
\textit{Buffer}     &    \textbf{Method}        &\textit{Class-IL}&\textit{Task-IL}  \\
\midrule

\multirow{4}{*}{50}       & ER~\cite{riemer2018learning}                  & 22.58 ± 0.71  &  66.28 ± 1.83 \\
                          & DER~\cite{buzzega2020dark}                & 29.35 ± 2.16  &  70.88 ± 0.83 \\
                          & DER++~\cite{buzzega2020dark}               & 31.92 ± 2.15  &  69.86 ± 1.96 \\
                          & \textbf{Ours}       & \textbf{37.24 ± 2.76}  &  \textbf{71.84 ± 1.82} \\
\midrule

\multirow{4}{*}{200}      & ER~\cite{riemer2018learning}                  & 36.22 ± 1.06  &  77.82 ± 1.24  \\
                          & DER~\cite{buzzega2020dark}                 & 46.18 ± 1.44  &  81.12 ± 1.59  \\
                          & DER++~\cite{buzzega2020dark}               & 51.84 ± 1.32  &  \textbf{82.66 ± 1.60}  \\
                          & \textbf{Ours}       & \textbf{52.83 ± 0.83} &  78.48 ± 1.42  \\
\midrule

\multirow{4}{*}{500}      & ER~\cite{riemer2018learning}                  & 49.70 ± 0.71  &  84.35 ± 0.79 \\
                          & DER~\cite{buzzega2020dark}                 & 56.58 ± 2.44  &  84.56 ± 1.19 \\
                          & DER++~\cite{buzzega2020dark}               & \textbf{60.28 ± 2.31}  &  \textbf{85.10 ± 0.93} \\
                          & \textbf{Ours}       & 59.36 ± 0.81  &  79.95 ± 0.61 \\
\midrule

\multirow{4}{*}{5120}     & ER~\cite{riemer2018learning}                  &  70.40 ± 1.30 &  89.20 ± 0.01  \\
                          & DER~\cite{buzzega2020dark}                 &  68.75 ± 0.25 &  89.35 ± 0.85   \\
                          & DER++~\cite{buzzega2020dark}               & \textbf{74.98 ± 0.66}  &  \textbf{90.72 ± 0.65} \\
                          & \textbf{Ours}       & 71.65 ± 0.82  &  85.93 ± 1.29 \\

\bottomrule
\end{tabular}
    \label{tab:microImagener}
\end{table}
Table~\ref{tab:seqCIFAR10} and Table~\ref{tab:microImagener} report results on Split-CIFAR-10 and Split-Micro-ImageNet, respectively. Our method yields the best Class-IL performance when tested with small/medium buffer size. It is also noteworthy that as the performance gain of our approach increases as the size of the buffer decreases. 
When buffer size becomes significantly larger, vanilla DER++ still achieves the best results, showing that retaining and replaying enough data (5,120 samples represent more than 10\% of the entire training set of CIFAR-10) is still the best option to alleviate catastrophic forgetting, although this goes in stark contrast to generalizing continual learning methods to real-world problems.
A similar behavior can be observed on the simpler Task-IL setting, where our method obtains the highest performance or is on par with existing methods under low--data availability regimes.

We also compare our approach to Co$^2$L~\cite{cha2021co2l}, that recently achieved state-the-art performance in both settings\footnote{Co$^2$L is not reported in Tables~\ref{tab:seqCIFAR10} and \ref{tab:microImagener}, as its training strategies are significantly different from the methods shown in those tables.}. 
Nevertheless, on Split-CIFAR-10, our method yields better Class-IL performance than Co$^2$L~\cite{cha2021co2l}, that reaches 65.57$\pm$1.37 with buffer size of 200 and  74.26$\pm$0.77 with buffer size of 500, respectively compared 70.86$\pm$0.95 and 75.07$\pm$0.41 by our method. The lower standard deviation also shows that our approach tends to be more stable across tasks.

\begin{figure}[t]
\centering
\includegraphics[width=1\linewidth]{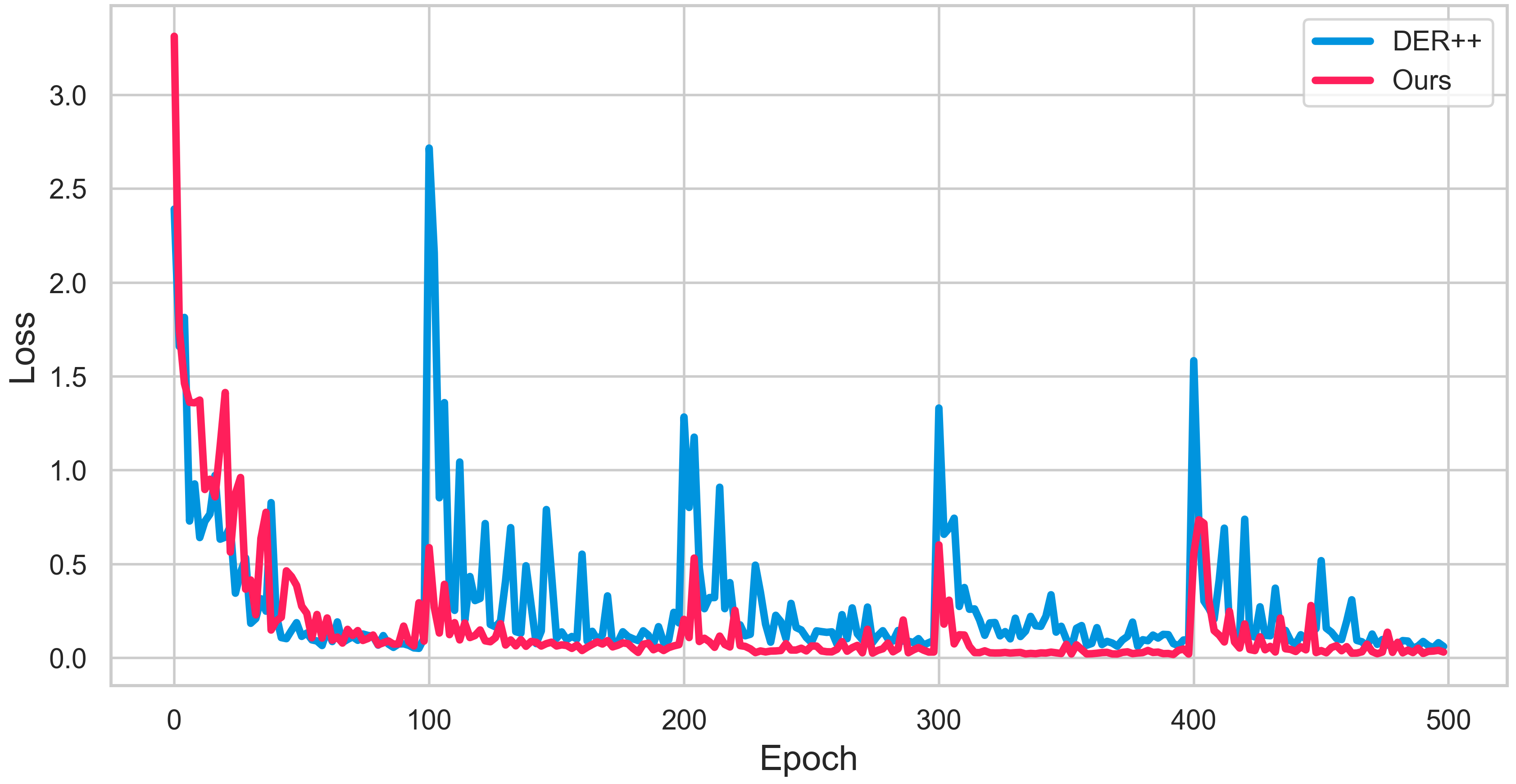}
\caption{Training loss trend for our approach (red) and DER++ (blue) on Split-Micro-ImageNet. When the model switches to a new task, MAH reduces loss peaks by assigning new classes to the most suitable available heads trained on auxiliary data.}
\label{fig:loss}
\end{figure}

Finally, we monitor the loss over consecutive tasks in order to evaluate the impact of auxiliary data during training.   
Fig.~\ref{fig:loss} shows the average training loss for vanilla DER++ and our method on Split-Micro-ImageNet. It can be observed that, as new tasks come in (every 100 epochs), the proposed approach shows a smoother loss surface, conversely to the vanilla counterpart that, instead, exhibits more noticeable peaks.
Thus, the proposed strategy also improves forward transfer and prevents disrupting gradient peaks when the model switches to new tasks, resembling non-continual learning scenarios.

\subsection{Ablation study}
In order to substantiate our design choices, we perform an ablation study to quantify the contribution of a) using the auxiliary data stream and b) the MAH strategy. 
The obtained results are reported in Table~\ref{tab:ablation} and compared to vanilla DER++ and DER++ with auxiliary data but \textbf{without MAH}. When the MAH strategy is not used, the classification heads are selected in a sequential order without making use of neural activation mapping between past (auxiliary) and current classes.

Results show that training with auxiliary data yields significant performance gains for all buffer sizes but 5,120, where replayed knowledge becomes prevalent. In all cases, MAH outperforms sequential head mapping.

\begin{table}
\centering
\caption{Ablation Study. Results obtained by the vanilla DER++ (first row of each block), DER++ with auxiliary data (second row), and the proposed method (third row), combining DER++ with auxiliary data and MAH strategy, for different buffer sizes.}
\label{tab:ablation}
\begin{tabular}{clll} 
\toprule
\textit{Buffer}           &    \textbf{Method}                              & \multicolumn{1}{c}{\textit{Class-IL}} & \multicolumn{1}{c}{\textit{Task-IL}}  \\ 
\midrule
\multirow{3}{*}{50}   & DER++                            & 49.28 ± 3.16                 & 86.14 ± 2.56                 \\
                      & \hspace{0.2 cm }$\hookrightarrow$  + AUX                & 52.74 ± 1.02                 & 88.51 ± 2.01                 \\
                      & \hspace{0.4 cm }$\hookrightarrow$ + MAH & \textbf{56.33 ± 0.95}        & \textbf{89.57 ± 2.47}        \\ 
\midrule
\multirow{3}{*}{200}  & DER++                            & 64.88 ± 1.17                 & 91.92 ± 0.60                 \\
                       & \hspace{0.2 cm }$\hookrightarrow$  + AUX                 & 69.91 ± 1.48                 & 92.57 ± 1.02                 \\
                      & \hspace{0.4 cm }$\hookrightarrow$ + MAH & \textbf{70.86 ± 0.95}        & \textbf{93.30 ± 0.64}        \\ 
\midrule
\multirow{3}{*}{500}  & DER++                                   & 72.70 ± 1.36                 & 93.88 ± 0.50                 \\
                      & \hspace{0.2 cm }$\hookrightarrow$ + AUX & 74.24 ± 0.61                 & \textbf{93.93 ± 0.47}        \\
                      & \hspace{0.4 cm }$\hookrightarrow$ + MAH & \textbf{75.07 ± 0.41}        & 93.62 ± 0.58       \\ 
\midrule
\multirow{3}{*}{5120} & DER++                            & \textbf{85.24 ± 0.49}        & \textbf{96.12 ± 0.21}        \\
                      & \hspace{0.2 cm }$\hookrightarrow$  + AUX                & 84.26 ± 0.22                 & 95.58 ± 0.22                 \\
                      & \hspace{0.4 cm }$\hookrightarrow$ + MAH & 84.56 ± 0.55                 & 95.84 ± 0.42                 \\
\bottomrule
\end{tabular}
\end{table}

\subsection{Effect of Pre-training}

We further investigate whether it is better to employ a backbone pre-trained on auxiliary data or to train it from scratch using the proposed strategy. The results of this analysis are reported in Table~\ref{tab:pretraining}: pre-training on auxiliary data appears to be always beneficial compared to training from scratch in DER++ and as the buffer size increases. On the contrary, on small buffers, using auxiliary data with our approach on a model trained from scratch yields better performance than pre-training. Furthermore, with our method, pre-training on auxiliary data leads instead to lower performance than training from scratch, showing that pre-training is not always a reliable alternative to continuously training with auxiliary data.
\begin{table}
\centering
\caption{Effect of Pre-training. Results obtained by the vanilla DER++ (first row of each block), DER++ pre-trained with auxiliary data (second row), the proposed method trained from scratch (third row), and the proposed model pre-trained with auxiliary data (fourth row), for different buffer sizes.}
\label{tab:pretraining}
\begin{tabular}{clll} 
\toprule
\textit{Buffer}           & \multicolumn{1}{c}{\textbf{Method}}             & \multicolumn{1}{c}{\textit{Class-IL}} & \multicolumn{1}{c}{\textit{Task-IL}}  \\ 
\midrule
\multirow{3}{*}{50}   & DER++                            & 49.28 ± 3.16                 & 86.14 ± 2.56                 \\
                      & DER++ with pre-training             & 50.82 ± 3.34                 & 84.02 ± 2.98                 \\
                      & Ours                 & \textbf{56.33 ± 0.95}        & \textbf{89.57 ± 2.47}        \\
                      & Ours with pre-training  & 53.52 ± 3.60        & 89.52 ± 0.88        \\ 
\midrule
\multirow{3}{*}{200}  & DER++                            & 64.88 ± 1.17                 & 91.92 ± 0.60                 \\
                      & DER++ with pre-training            & 68.71 ± 1.01        & 92.43 ± 0.53       \\
                      & Ours                 & \textbf{70.86 ± 0.95}        & \textbf{93.30 ± 0.64}        \\
                      & Ours with pre-training & 65.17 ± 2.67                 & 91.35 ± 1.71                 \\ 
\midrule
\multirow{3}{*}{500}  & DER++                            & 72.70 ± 1.36                 & 93.88 ± 0.50                 \\
                      & DER++ with pre-training            & \textbf{75.91 ± 0.26}        & \textbf{94.39 ± 0.29}        \\
                      & Ours                 & 75.07 ± 0.41                 & 93.62 ± 0.58                 \\
                      & Ours with pre-training & 71.39 ± 2.77                 & 91.77 ± 0.76                 \\ 
\midrule
\multirow{3}{*}{5120} & DER++                            & 85.24 ± 0.49       & 96.12 ± 0.21   \\
                      & DER++ with pre-training            & \textbf{86.60 ± 0.42}        & \textbf{96.29 ± 0.09}        \\
                      & Ours                 & 84.56 ± 0.55                 & 95.84 ± 0.42                 \\
                      & Ours with pre-training & 82.20 ± 1.37                 & 94.65 ± 0.36                 \\
\bottomrule
\end{tabular}
\end{table}

\subsection{Generative Auxiliary Model}

In the previous sections, we have consistently observed that using auxiliary data helps retaining knowledge of previous tasks, especially with limited buffer size. However, it is not efficient to maintain the auxiliary data in memory, as in that case it is still preferable to simply use a larger buffer. 

A viable alternative would be to replace the auxiliary stream with a generative replay model and use generated samples during task training.
In order to investigate the feasibility of this option, we use a generative adversarial network (GAN)~\cite{goodfellow2014generative} to learn the distribution of auxiliary data, and employ synthetic images in place of real ones in Eq.~\ref{eq:aux} in our method.

We then replicate the experiments carried out on Split-CIFAR-10 by pre-training a BigGAN model~\cite{brock2019large} on super-classes of CIFAR-100 used as auxiliary data, and compare the results with those obtained when using real images. As it can be seen in Table~\ref{tab:CIFAR10_gan}, performance achieved when using generated images follow the same behavior observed with real data, i.e., performance  increase in settings with small buffer size, while the approach is less beneficial when the replay memory increases.

\begin{table}[]
    \centering
\caption{Effect of replacing auxiliary data with a GAN. Results obtained on Split-CIFAR-10 when using no auxiliary data (vanilla DER++), real auxiliary data (the proposed method) and synthetic data.}
\begin{tabular}{clcc}
\toprule
\textit{Buffer}           &    \textbf{Aux. data}      &     \textit{Class-IL}    &   \textit{Task-IL}  \\
\midrule

\multirow{3}{*}{50}       & \textit{none}      & 49.28 ± 3.16  &  86.14 ± 2.56 \\
                          & real       & 56.33 ± 0.95  &  89.57 ± 2.47 \\
                          & synthetic        & 54.47 ± 3.05  &  89.23 ± 1.83 \\
\midrule

\multirow{3}{*}{200}    & \textit{none}      & 64.88 ± 1.17  &  91.92 ± 0.60 \\
                        & real       & 70.86 ± 0.95  &  93.30 ± 0.64 \\
                        & synthetic        & 68.84 ± 0.77  &  92.85 ± 0.24 \\
                          
\midrule

\multirow{3}{*}{500}    & \textit{none}      & 72.70 ± 1.36  &  93.88 ± 0.50 \\
                        & real       & 75.07 ± 0.41  &  93.62 ± 0.58 \\
                        & synthetic        & 74.35 ± 1.04  &  93.42 ± 0.40 \\
                          
\midrule

\multirow{3}{*}{5120}   & \textit{none}     & 85.24 ± 0.49  &  96.12 ± 0.21 \\
                        & real      & 84.56 ± 0.55  &  95.84 ± 0.42 \\
                        & synthetic       & 84.14 ± 0.45  &  95.65 ± 0.23 \\

\bottomrule
\end{tabular}
    \label{tab:CIFAR10_gan}
\end{table}

\section{Conclusion}
In this work, we propose a novel approach for improving continual learning accuracy by leveraging external data. Our experiments show that providing the model with an additional auxiliary stream leads to an increase in performance, especially when the employed memory buffer is small, and to more stable training at the beginning of each task. 
We also observe that our approach works consistently better, on small buffer settings, than alternative knowledge transfer strategies such as direct pre-training on auxiliary data. The approximation of the auxiliary data distribution through the use of generative models also outperforms state-of-the-art models, thus indicating the future direction of this work, i.e., a more effective modeling of previous real-world knowledge as it seems to happen in the human hippocampus~\cite{pmid27941790}.

\section*{Acknowledgment}
This research has been partially supported by: \emph{RehaStart} project, Regione Sicilia (PO FESR 2014/2020, Azione 1.1.5, N. 08ME6201000222, CUP G79J18000610007); Piano della Ricerca di Ateneo 2020/2022, PIACERI Starting Grant Universit\`a di Catania, Project: \emph{BrAIn: Adaptive Brain-Derived
Artificial Intelligence Methods for Event Detection};  Piano della Ricerca di Ateneo 2020/2022, Linea 2D, Universit\`a di Catania.


\begin{thebibliography}{10}

\providecommand{\url}[1]{#1}
\csname url@samestyle\endcsname
\providecommand{\newblock}{\relax}
\providecommand{\bibinfo}[2]{#2}
\providecommand{\BIBentrySTDinterwordspacing}{\spaceskip=0pt\relax}
\providecommand{\BIBentryALTinterwordstretchfactor}{4}
\providecommand{\BIBentryALTinterwordspacing}{\spaceskip=\fontdimen2\font plus
\BIBentryALTinterwordstretchfactor\fontdimen3\font minus \fontdimen4\font\relax}
\providecommand{\BIBforeignlanguage}[2]{{%
\expandafter\ifx\csname l@#1\endcsname\relax
\typeout{** WARNING: IEEEtran.bst: No hyphenation pattern has been}%
\typeout{** loaded for the language `#1'. Using the pattern for}%
\typeout{** the default language instead.}%
\else
\language=\csname l@#1\endcsname
\fi
#2}}
\providecommand{\BIBdecl}{\relax}
\BIBdecl

\bibitem{cichon2015branch}
J.~Cichon and W.-B. Gan, ``Branch-specific dendritic ca 2+ spikes cause
  persistent synaptic plasticity,'' \emph{Nature}, 2015.

\bibitem{mccloskey1989catastrophic}
M.~McCloskey and N.~J. Cohen, ``Catastrophic interference in connectionist
  networks: The sequential learning problem,'' in \emph{Psychol. Learn. Motiv.}, 1989.

\bibitem{ratcliff1990connectionist}
R.~Ratcliff, ``{Connectionist models of recognition memory: constraints imposed
  by learning and forgetting functions.}'' \emph{Psychol. Rev.}, 1990.

\bibitem{parisi2019continual}
G.~I. Parisi, R.~Kemker, J.~L. Part, C.~Kanan, and S.~Wermter, ``Continual
  lifelong learning with neural networks: A review,'' \emph{Neural Netw.},
  2019.

\bibitem{delange2021continual}
M.~D. Lange, R.~Aljundi, M.~Masana, S.~Parisot, X.~Jia, A.~Leonardis, G.~G.
  Slabaugh, and T.~Tuytelaars, ``Continual learning: A comparative study on how
  to defy forgetting in classification tasks,'' \emph{IEEE TPAMI}, 2021.

\bibitem{kirkpatrick2017overcoming}
J.~Kirkpatrick, R.~Pascanu, N.~Rabinowitz, J.~Veness, G.~Desjardins, A.~A.
  Rusu, K.~Milan, J.~Quan, T.~Ramalho, A.~Grabska-Barwinska \emph{et~al.},
  ``{Overcoming catastrophic forgetting in neural networks},'' \emph{PNAS},
  2017.

\bibitem{li2017learning}
Z.~Li and D.~Hoiem, ``{Learning without forgetting},'' \emph{IEEE TPAMI}, 2017.

\bibitem{rusu2016progressive}
A.~A. Rusu, N.~C. Rabinowitz, G.~Desjardins, H.~Soyer, J.~Kirkpatrick,
  K.~Kavukcuoglu, R.~Pascanu, and R.~Hadsell, ``Progressive neural networks,''
  \emph{arXiv preprint arXiv:1606.04671}, 2016.

\bibitem{serra2018overcoming}
J.~Serra, D.~Suris, M.~Miron, and A.~Karatzoglou, ``Overcoming catastrophic
  forgetting with hard attention to the task,'' in \emph{ICML}, 2018.

\bibitem{chaudhry2019tiny}
A.~Chaudhry, M.~Rohrbach, M.~Elhoseiny, T.~Ajanthan, P.~K. Dokania, P.~H. Torr,
  and M.~Ranzato, ``{On tiny episodic memories in continual learning},'' in
  \emph{ICML Workshops}, 2019.

\bibitem{shin2017continual}
H.~Shin, J.~K. Lee, J.~Kim, and J.~Kim, ``Continual learning with deep
  generative replay,'' in \emph{ANeurIPS}, 2017.

\bibitem{french1991using}
R.~M. French, ``{Using semi-distributed representations to overcome
  catastrophic forgetting in connectionist networks},'' in \emph{Proc. Annu.
  Conf. Cogn. Sci. Soc.}, 1991.

\bibitem{mcrae1993catastrophic}
K.~McRae and P.~A. Hetherington, ``Catastrophic interference is eliminated in
  pretrained networks,'' in \emph{Proc. Annu. Conf. Cogn. Sci. Soc.}, 1993.

\bibitem{hendrycks2019using}
D.~Hendrycks, K.~Lee, and M.~Mazeika, ``Using pre-training can improve model
  robustness and uncertainty,'' in \emph{ICML}, 2019.

\bibitem{ddf607f82ffd4ff4a2e7e9e4d833a78d}
S.~Sakellaridi, V.~Christopoulos, T.~Aflalo, K.~Pejsa, E.~Rosario,
  D.~Ouellette, N.~Pouratian, and R.~Andersen, ``Intrinsic variable learning
  for brain-machine interface control by human anterior intraparietal cortex,''
  \emph{Neuron}, 2019.

\bibitem{silver2016mastering}
D.~Silver, A.~Huang, C.~J. Maddison, A.~Guez, L.~Sifre, G.~Van Den~Driessche,
  J.~Schrittwieser, I.~Antonoglou, V.~Panneershelvam, M.~Lanctot \emph{et~al.},
  ``Mastering the game of go with deep neural networks and tree search,''
  \emph{Nature}, 2016.

\bibitem{vinyals2019grandmaster}
O.~Vinyals, I.~Babuschkin, W.~M. Czarnecki, M.~Mathieu, A.~Dudzik, J.~Chung,
  D.~H. Choi, R.~Powell, T.~Ewalds, P.~Georgiev \emph{et~al.}, ``Grandmaster
  level in starcraft ii using multi-agent reinforcement learning,''
  \emph{Nature}, 2019.

\bibitem{lewandowsky1995catastrophic}
S.~Lewandowsky and S.-C. Li, ``Catastrophic interference in neural networks:
  Causes, solutions, and data,'' in \emph{Interference and inhibition in
  cognition}.\hskip 1em plus 0.5em minus 0.4em\relax Elsevier, 1995.

\bibitem{chaudhry2020continual}
A.~Chaudhry, N.~Khan, P.~K. Dokania, and P.~H. Torr, ``Continual learning in
  low-rank orthogonal subspaces,'' in \emph{ANeurIPS}, 2020.

\bibitem{mallya2018packnet}
A.~Mallya and S.~Lazebnik, ``Packnet: Adding multiple tasks to a single network
  by iterative pruning,'' in \emph{CVPR}, 2018.

\bibitem{jung2020continual}
S.~Jung, H.~Ahn, S.~Cha, and T.~Moon, ``Continual learning with node-importance
  based adaptive group sparse regularization,'' in \emph{ANeurIPS}, 2020.

\bibitem{rajasegaran2019random}
J.~Rajasegaran, M.~Hayat, S.~H. Khan, F.~S. Khan, and L.~Shao, ``Random path
  selection for continual learning,'' in \emph{ANeurIPS}, 2019.

\bibitem{goodfellow2014empirical}
I.~J. Goodfellow, M.~Mirza, X.~Da, A.~C. Courville, and Y.~Bengio, ``An
  empirical investigation of catastrophic forgeting in gradient-based neural
  networks,'' \emph{ICLR}, 2014.

\bibitem{mirzadeh2020understanding}
S.~I. Mirzadeh, M.~Farajtabar, R.~Pascanu, and H.~Ghasemzadeh, ``Understanding
  the role of training regimes in continual learning,'' in \emph{ANeurIPS},
  2020.

\bibitem{lopez2017gradient}
D.~Lopez-Paz and M.~Ranzato, ``{Gradient episodic memory for continual
  learning},'' in \emph{ANeurIPS}, 2017.

\bibitem{chaudhry2018efficient}
A.~Chaudhry, M.~Ranzato, M.~Rohrbach, and M.~Elhoseiny, ``{Efficient Lifelong
  Learning with A-GEM},'' in \emph{ICLR}, 2019.

\bibitem{farajtabar2020orthogonal}
M.~Farajtabar, N.~Azizan, A.~Mott, and A.~Li, ``Orthogonal gradient descent for
  continual learning,'' in \emph{AISTATS}, 2020.

\bibitem{buzzega2020dark}
P.~Buzzega, M.~Boschini, A.~Porrello, D.~Abati, and S.~Calderara, ``{Dark
  Experience for General Continual Learning: a Strong, Simple Baseline},'' in
  \emph{ANeurIPS}, 2020.

\bibitem{aljundi2019gradient}
R.~Aljundi, M.~Lin, B.~Goujaud, and Y.~Bengio, ``{Gradient based sample
  selection for online continual learning},'' in \emph{ANeurIPS}, 2019.

\bibitem{aljundi2019online}
R.~Aljundi, E.~Belilovsky, T.~Tuytelaars, L.~Charlin, M.~Caccia, M.~Lin, and
  L.~Page-Caccia, ``{Online continual learning with maximal interfered
  retrieval},'' in \emph{ANeurIPS}, 2019.

\bibitem{riemer2019scalable}
M.~Riemer, T.~Klinger, D.~Bouneffouf, and M.~Franceschini, ``Scalable
  recollections for continual lifelong learning,'' in \emph{AAAI}, 2019.

\bibitem{rebuffi2017icarl}
S.-A. Rebuffi, A.~Kolesnikov, G.~Sperl, and C.~H. Lampert, ``{icarl:
  Incremental classifier and representation learning},'' in \emph{CVPR}, 2017.

\bibitem{hou2019learning}
S.~Hou, X.~Pan, C.~C. Loy, Z.~Wang, and D.~Lin, ``{Learning a unified
  classifier incrementally via rebalancing},'' in \emph{CVPR}, 2019.

\bibitem{wu2019large}
Y.~Wu, Y.~Chen, L.~Wang, Y.~Ye, Z.~Liu, Y.~Guo, and Y.~Fu, ``{Large scale
  incremental learning},'' in \emph{CVPR}, 2019.

\bibitem{cha2021co2l}
H.~Cha, J.~Lee, and J.~Shin, ``Co2l: Contrastive continual learning,'' in
  \emph{ICCV}, 2021.

\bibitem{pham2021dualnet}
Q.~Pham, C.~Liu, and S.~Hoi, ``Dualnet: Continual learning, fast and slow,'' in
  \emph{ANeurIPS}, 2021.

\bibitem{kim2021continual}
C.~D. Kim, J.~Jeong, S.~Moon, and G.~Kim, ``Continual learning on noisy data
  streams via self-purified replay,'' in \emph{ICCV}, 2021.

\bibitem{prabhu2020gdumb}
A.~Prabhu, P.~H. Torr, and P.~K. Dokania, ``{GDumb: A simple approach that
  questions our progress in continual learning},'' in \emph{ECCV}, 2020.

\bibitem{farquhar2018towards}
S.~Farquhar and Y.~Gal, ``{Towards Robust Evaluations of Continual Learning},''
  in \emph{ICML Workshops}, 2018.

\bibitem{van2019three}
G.~M. van~de Ven and A.~S. Tolias, ``{Three continual learning scenarios},'' in
  \emph{NeurIPS Workshops}, 2018.

\bibitem{riemer2018learning}
M.~Riemer, I.~Cases, R.~Ajemian, M.~Liu, I.~Rish, Y.~Tu, and G.~Tesauro,
  ``{Learning to Learn without Forgetting by Maximizing Transfer and Minimizing
  Interference},'' in \emph{ICLR}, 2019.

\bibitem{benjamin2018measuring}
A.~S. Benjamin, D.~Rolnick, and K.~Kording, ``Measuring and regularizing
  networks in function space,'' in \emph{ICLR}, 2019.

\bibitem{chaudhry2020using}
A.~Chaudhry, A.~Gordo, P.~K. Dokania, P.~Torr, and D.~Lopez-Paz, ``Using
  hindsight to anchor past knowledge in continual learning,'' in \emph{AAAI},
  2021.

\bibitem{zenke2017continual}
F.~Zenke, B.~Poole, and S.~Ganguli, ``Continual learning through synaptic
  intelligence,'' in \emph{ICML}, 2017.

\bibitem{Le2015TinyIV}
Stanford, ``{Tiny ImageNet Challenge (CS231n)},'' 2015,
  \url{https://www.kaggle.com/c/tiny-imagenet}.

\bibitem{he2016deep}
K.~He, X.~Zhang, S.~Ren, and J.~Sun, ``Deep residual learning for image
  recognition,'' in \emph{ICPR}, 2016.

\bibitem{goodfellow2014generative}
I.~Goodfellow, J.~Pouget-Abadie, M.~Mirza, B.~Xu, D.~Warde-Farley, S.~Ozair,
  A.~Courville, and Y.~Bengio, ``Generative adversarial nets,'' in
  \emph{ANeurIPS}, 2014.

\bibitem{brock2019large}
A.~Brock, J.~Donahue, and K.~Simonyan, ``Large scale gan training for high
  fidelity natural image synthesis,'' in \emph{ICLR}, 2019.

\bibitem{pmid27941790}
G.~Rothschild, E.~Eban, and L.~M. Frank, ``{{A} cortical-hippocampal-cortical
  loop of information processing during memory consolidation},'' \emph{Nat.Neurosci.}, 2017.

\end{thebibliography}
\end{document}